\documentclass[runningheads]{llncs}

\usepackage{eccv}

\usepackage{eccvabbrv}

\usepackage{graphicx}
\usepackage{booktabs}

\usepackage[accsupp]{axessibility}  

\usepackage{booktabs}       
\usepackage{amsfonts}       
\usepackage{nicefrac}       
\usepackage{microtype}      
\usepackage{xcolor}         
\usepackage{enumitem}
\usepackage{graphicx}
\usepackage{amsmath}
\usepackage{changepage}
\usepackage{listings}
\usepackage{subcaption}
\usepackage{multirow}
\usepackage{array}
\usepackage{color,colortbl}
\definecolor{Gray}{gray}{0.95}
\usepackage{wrapfig}
\usepackage{floatrow}
\newfloatcommand{capbtabbox}{table}[][\FBwidth]
\usepackage{mathtools}
\usepackage{pifont}

\newcommand{\cmark}{\ding{51}}%
\usepackage{refcount}

\usepackage{xr}
\usepackage{hyperref}

\usepackage{orcidlink}

\begin{document}

\title{Prompting Language-Informed Distribution for Compositional Zero-Shot Learning} 

\titlerunning{PLID for CZSL}

\author{Wentao Bao\inst{1}\orcidlink{0000-0003-2571-3341} \and
Lichang Chen\inst{2}\orcidlink{0000-0003-1384-2760} \and
Heng Huang\inst{2}\orcidlink{0000-0002-3483-8333} \and 
Yu Kong\inst{1}\orcidlink{0000-0001-6271-4082}}

\authorrunning{W. Bao et al.}

\institute{Department of Computer Science and Engineering, Michigan State University 
\and Department of Computer Science, University of Maryland \\
\email{\{baowenta,yukong\}@msu.edu}, \email{\{bobchen,heng\}@umd.edu}}

\maketitle

\begin{abstract}
  Compositional zero-shot learning (CZSL) task aims to recognize unseen compositional visual concepts, \eg, \texttt{sliced tomatoes}, where the model is learned only from the seen compositions, \eg, \texttt{sliced potatoes} and \texttt{red tomatoes}. Thanks to the prompt tuning on large pre-trained visual language models such as CLIP, recent literature shows impressively better CZSL performance than traditional vision-based methods. However, the key aspects that impact the generalization to unseen compositions, including the \emph{diversity} and \emph{informativeness} of class context, and the \emph{entanglement} between visual primitives, \ie, state and object, are not properly addressed in existing CLIP-based CZSL literature. In this paper, we propose a model by prompting the language-informed distribution, aka., $\mathbb{PLID}$, for the CZSL task. Specifically, the $\mathbb{PLID}$ leverages pre-trained large language models (LLM) to (\textit{i}) formulate the language-informed class distributions which are diverse and informative, and (\textit{ii}) enhance the compositionality of the class embedding. Moreover, a visual-language primitive decomposition (VLPD) module is proposed to dynamically fuse the classification decisions from the compositional and the primitive space. Orthogonal to the existing literature of soft, hard, or distributional prompts, our method advocates prompting the LLM-supported class distributions, leading to a better zero-shot generalization. Experimental results on MIT-States, UT-Zappos, and C-GQA datasets show the superior performance of the $\mathbb{PLID}$ to the prior arts. Our code and models are released: \url{https://github.com/Cogito2012/PLID}.
\end{abstract}

\section{Introduction}
\label{sec:intro}

Compositional visual recognition is a fundamental characteristic of human intelligence~\cite{lake2017building} but it is challenging for modern deep learning systems. For example, humans can easily recognize unseen \texttt{sliced tomatoes} after seeing \texttt{sliced potatoes} and \texttt{red tomatoes}. Such a compositional zero-shot learning (CZSL) capability is valuable in that, novel visual concepts from a huge combinatorial semantic space could be recognized without ``seeing'' any of their training data. For example, the C-GQA~\cite{naeem2021learning} dataset contains 413 states and 674 objects. This implies a total of at least 278K compositional classes in an open world while only 2\% of them are accessible in training. Therefore, CZSL can significantly reduce the need for large-scale training data.

Traditional vision-based methods either directly learn the visual feature of compositions, or try to first decompose the visual data into representations of simple primitives, \ie, states and objects, and then learn to re-compose the compositions~\cite{misra2017red,atzmon2020causal,zou2020compositional,huynh2020compositional,karthik2022kg,tokmakov2019learning,naeem2021learning,zhang2022learning,mancini2021open,li2022siamese}. Thanks to the recent large pre-trained vision-language models (VLM) such as CLIP~\cite{CLIP}, state-of-the-art CZSL methods have been developed~\cite{csp_iclr23,dfsp_cvpr23,xu2022prompting,troika_arxiv23}. For instance, CSP~\cite{csp_iclr23} inherits the hard prompt template of the CLIP, \ie, \emph{a photo of} $[\texttt{state}][\texttt{object}]$ where only the embeddings of the states and objects are trained. The following methods~\cite{dfsp_cvpr23,xu2022prompting,troika_arxiv23} use soft prompt introduced in CoOp~\cite{zhou2022coop}, where the embeddings of the prompt template are jointly optimized, leading to a better CZSL performance. The impressive performance of CLIP-based CZSL methods benefits from the sufficiently good feature alignment between the image and text modalities, and the prompting techniques for adapting the aligned features to recognizing compositional classes. 

\begin{figure}[t]
    \centering
    \includegraphics[width=\linewidth]{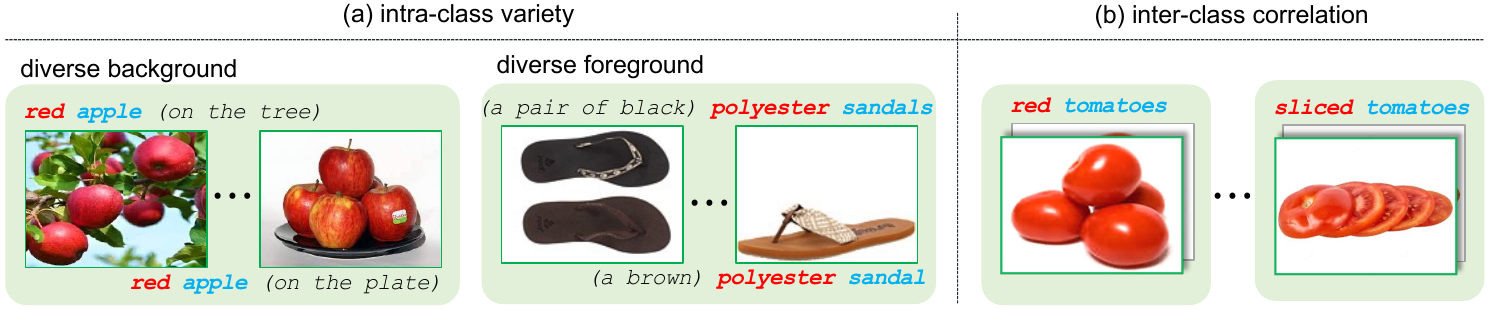}
    \captionsetup{font=small,belowskip=0pt}
    \caption{\textbf{Challenges of compositional recognition.} \textbf{(a)} images of the same compositional class appear differently due to diverse visual backgrounds or foregrounds. \textbf{(b)} \texttt{red tomatoes} and \texttt{sliced tomatoes} are visually correlated because 1) both are \texttt{tomatoes} object, and 2) the object \texttt{tomatoes} is inherently entangled with the state \texttt{red}, resulting in the need of primitive decomposition.}
    \label{fig:class}
\end{figure}

Despite the success of existing CLIP-based methods, we find several key considerations to prompt the pre-trained CLIP for better CZSL modeling. First, the \emph{diversity} and \emph{informativeness} of prompts are both important to distinguish between compositional classes. CZSL can be treated as zero-shot learning on fine-grained categories, which requires a fine-grained context to prompt the CLIP model~\cite{CLIP,lu2022prompt}. However, to contextualize a class with fine granularity, the hard prompt in \cite{CLIP} suffers from the heuristic design of prompt templates, and a single prompt for each class lacks diversity to capture the intra-class variance of visual data (\cref{fig:class}a). Though the ProDA~\cite{lu2022prompt} proposes to learn a collection of prompts that formulate class-specific distribution to address the diversity, the lack of~\emph{language informativeness} in their prompts limits their performance on fine-grained compositional categories. Second, the entanglement between visual primitives, \eg\texttt{red} and~\texttt{tomatoes} in \cref{fig:class}b, incurs difficulty in learning decomposable visual representations that are useful for compositional generalization~\cite{liu2022simple,karthik2022kg}, while such a capability is missing in~\cite{csp_iclr23,xu2022prompting}. Though the more recent work~\cite{dfsp_cvpr23,troika_arxiv23} learn to decompose the primitives and considers the re-composed compositional predictions, their language-only decomposition and probability-level mixup potentially limit the generalizability in the open-world.

In this paper, we propose a novel CLIP-based method for the CZSL task by prompting the language-informed distributions ($\mathbb{PLID}$) over both the compositional and primitive categories. To learn the diverse and informative textual class representations, the $\mathbb{PLID}$ leverages off-the-shelf large language models (LLM) to build the class-specific distributions and to enhance the class embeddings. 
Furthermore, we propose a visual language primitive decomposition (VLPD) module to decompose the image data into simple primitives for recognition of state and objects. Eventually, the compositional classification is performed by fusing the decisions from both the compositional and primitive spaces. The proposed $\mathbb{PLID}$ shows state-of-the-art performance on CZSL benchmarks such as MIT-States~\cite{mitstates}, UT-Zappos~\cite{utzappos}, and C-GQA~\cite{naeem2021learning}.

Note that our method is orthogonal to the existing hard prompt~\cite{CLIP}, soft prompt tuning~\cite{zhou2022coop}, and prompt distribution learning~\cite{lu2022prompt,kwon2023probabilistic,liu2023patch,derakhshani2023bayesian}. We advocate prompting the distribution of informative LLM-based class descriptions. From a classification perspective, this is grounded on the classification-by-description~\cite{menon2022visual,maniparambil2023enhancing,yan2023learning,he2022synthetic}, that LLM-generated text enables more informative class representations. Compared to the deterministic soft or hard prompt aforementioned, our distribution modeling could capture the intra-class diversity and inter-class correlation for better zero-shot generalization. Compared to the existing prompt distribution learning approaches, the class context is more linguistically interpretable and provides fine-grained descriptive information about the class. Our method is also parameter-efficient without the need to optimize a large collection of prompts. Specific to the CZSL task, the enhanced class embeddings by LLM descriptions enable visual language primitive decomposition and decision fusion in both compositional and primitive space, which eventually benefits the generalization to the unseen.

In summary, the contributions are as follows. (\textit{i}) We develop a $\mathbb{PLID}$ method that advocates prompting the language-informed distribution for compositional zero-shot learning, which is orthogonal to existing soft or hard prompting and distributional prompt learning. (\textit{ii}) We propose primitive decomposition with stochastic logit mixup to fuse the classification decision from compositional and primitive predictions. (\textit{iii}) We empirically show that $\mathbb{PLID}$ could achieve superior performance to prior arts in both the closed-world and open-world settings on MIT-States, UT-Zappos, and C-GQA datasets.

\section{Related Work}

\textbf{Prompt Learning in VLM.} Vision-Language Models (VLM) such as the CLIP~\cite{CLIP} pre-trained on web-scale datasets recently gained substantial attention for their strong zero-shot recognition capability on various downstream tasks. Such a capability is typically achieved by performing prompt engineering to adapt pre-trained VLMs. Early prompting technique such as the hard prompt in CLIP uses the heuristic template ``\emph{a photo of} \texttt{[CLS]}'' as the textual input. Recently, the soft prompt tuning method in CoOp~\cite{zhou2022coop}, CoCoOp~\cite{Zhou_2022_CVPR}, and ResPT~\cite{ResPrompt} that uses learnable embedding as the textual context of class names significantly improved the model adaptation performance. This technique is further utilized in MaPLe~\cite{khattak2023maple} that enables multi-modal prompt learning for both image and text. However, the prompts of these methods are deterministic and lack the diversity to capture the appearance variety in fine-grained visual data, so they are prone to overfitting the training data. To handle this issue, ProDA~\cite{lu2022prompt} explicitly introduces a collection of soft prompts to construct the class-specific Gaussian distribution, which results in better zero-shot performance and inspires the recent success of PPL~\cite{kwon2023probabilistic} in the dense prediction task. Similarly, the PBPrompt~\cite{liu2023patch} uses neural networks to predict the class-specific prompt distribution and utilizes optimal transport to align the stochastically sampled soft prompts and image patch tokens. The recent work~\cite{derakhshani2023bayesian} assumes the latent embedding of prompt input follows a Gaussian prior and adopts variational inference to learn the latent distribution. In this paper, in order to take the merits of the \emph{informativeness} of hard prompt and the \emph{diversity} of distributional modeling, we adopt the soft prompt to adapt the distributions supported by LLM-generated class descriptions.

\textbf{Compositional Zero-Shot Learning (CZSL).} For a long period, the CZSL task has been studied from a vision-based perspective in literature. They either directly learn the compositional visual features or disentangle the visual features into simple primitives, \ie, states and objects. For example,~\cite{nagarajan2018attributes,li2020symmetry,naeem2021learning} performs a direct classification by projecting the compositional visual features into a common feature space, and~\cite{lu2016visual,misra2017red,atzmon2020causal,huynh2020compositional,zou2020compositional,karthik2022kg,liu2022simple} decompose the visual feature into simple primitives so that the compositional recognition can be achieved by learning to recompose from the primitives. 
Though the recent large-scale pre-trained CLIP model shows impressive zero-shot capability, it is found to struggle to work well for compositional reasoning~\cite{ma2023crepe,yuksekgonul2022and,lewis2022does}.
Thanks to the recent prompt learning~\cite{zhou2022coop}, the CZSL task has been dominated by CLIP-based approaches~\cite{csp_iclr23,dfsp_cvpr23,xu2022prompting,troika_arxiv23,li2024context,zheng2024caila}. The common idea is to prompt the frozen CLIP model to separately learn the textual embeddings of simple primitives, which empirically show strong compositionality for zero-shot generalization. Different to~\cite{zheng2024caila,li2024context} that develop primitive adapters and~\cite{dfsp_cvpr23,xu2022prompting,troika_arxiv23} that use learnable prompts for deterministic vision-language alignment, our method takes the benefit of learnable prompt and LLM-generated text for distributional alignment, addressing the importance of diversity and informativeness for zero-shot generalization.

\section{Preliminaries}
\textbf{CZSL Task Formulation.} The CZSL task aims to recognize images of a compositional category $y\in \mathcal{C}$, where the semantic space $\mathcal{C}$ is a Cartesian product between the state space $\mathcal{S}=\{s_1,\ldots,s_{\lvert\mathcal{S}\rvert}\}$ and object space $\mathcal{O}=\{o_1,\ldots,o_{\lvert\mathcal{O}\rvert}\}$, \ie, $\mathcal{C}=\mathcal{S}\times\mathcal{O}$. For example, as shown in \cref{fig:class}, a model trained on images of \texttt{red apple} and \texttt{sliced tomatoes} needs to additionally recognize an image of \texttt{sliced apple}. In training, only a set of \textbf{seen} compositions is available. In closed-world testing, the model needs to recognize images from both the \textbf{seen} compositions in $\mathcal{C}^{(s)}$ and the \textbf{unseen} compositions in $\mathcal{C}^{(u)}$ that are assumed to be feasible, where the cardinality $\lvert \mathcal{C}^{(s)}\cup \mathcal{C}^{(u)} \rvert \ll \lvert \mathcal{C} \rvert$ since most of the compositions in $\mathcal{C}$ are practically not feasible. In open-world testing, the model needs to recognize images given any composition in $\mathcal{C}$. 

\textbf{VLMs for CZSL.} Large pre-trained VLMs such as CLIP~\cite{CLIP} have recently been utilized by CSP~\cite{csp_iclr23} for the CZSL task. The core idea of CSP is to represent the text embeddings of states in $\mathcal{S}$ and objects in $\mathcal{O}$ as learnable parameters and contextualize them with the hard prompt template ``\emph{a photo of} $[\texttt{s}] [\texttt{o}]$'' as the input of the CLIP text encoder, where $[\texttt{s}]\in \mathcal{S}$ and $[\texttt{o}]\in\mathcal{O}$. Given an image $\mathbf{x}$, by using the cosine similarity ($\texttt{cos}$) as the logit, the class probability of the composition $y$ is defined as $p_{\boldsymbol{\theta}}(y|\mathbf{x})=\texttt{softmax}(\texttt{cos}(\mathbf{v},\mathbf{t}_y))$,
where $\boldsymbol{\theta}$ are the $\lvert\mathcal{S}\rvert+\lvert\mathcal{O}\rvert$ learnable parameters, $\mathbf{v}$ and $\mathbf{t}_y$ are the image feature and class text embedding, respectively. 

In training, the prediction $p_{\boldsymbol{\theta}}(\hat{y}|\mathbf{x})$ is supervised by multi-class cross-entropy loss. In CZSL testing, a test image is recognized by finding the compositional class $c\in\mathcal{C}$ which has the maximum $\texttt{cos}(\mathbf{v},\mathbf{t}_c)$. The CSP method is simple, parameters efficient, and it largely outperforms traditional approaches. However, due to the lack of diversity and informativeness in prompting, the zero-shot capability of CLIP is not fully exploited by CSP for the CZSL task.

\section{Proposed Method}

\begin{figure}[t]
    \centering
    \includegraphics[width=\linewidth]{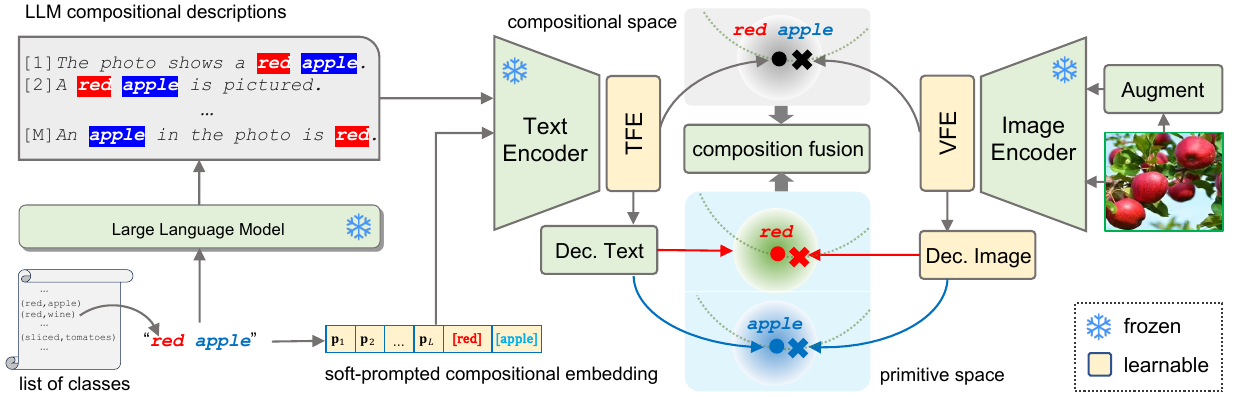}
    \caption{\footnotesize{Overview of $\mathbb{PLID}$. The model is developed for the CZSL task by aligning the semantics of image $\mathbf{x}$ (\eg, image on the right) and compositional class $y=(s,o)$ (\eg, ``red apple'') via a frozen CLIP~\cite{CLIP}. It constructs language-informed text distributions in both compositional and primitive (attribute and object) spaces (middle part) by soft prompting and LLM-generated class descriptions (left part). The features of the image and text are enhanced by text and visual feature enhancement (TFE and VFE).
    Eventually, the compositional decisions from the two spaces are fused as the prediction.
    }
    }
    \label{fig:framework}
\end{figure}

\textbf{Overview.} \Cref{fig:framework} shows an overview of the $\mathbb{PLID}$. The basic idea is to use LLMs to generate sentence-level descriptions for each compositional class and learn to prompt the class-wise text distributions (supported by the descriptions) to be aligned with image data. Besides, we introduce visual language primitive decomposition (VLPD) and stochastic logit mixup (SLM) to enable recognition at both compositional and primitive levels. In testing, an image is recognized by fusing the decisions from the directly predicted and the recomposed compositions.

\subsection{Prompting Language-Informed Distribution}

\textbf{Motivation.} To adapt the large pre-trained CLIP~\cite{CLIP} to downstream tasks, recent distributional prompt learning~\cite{lu2022prompt,kwon2023probabilistic,liu2023patch,derakhshani2023bayesian} shows the importance of \emph{context diversity} by distribution modeling for strong generalization. Motivated by the inherent fine-granularity of compositional recognition in the CZSL task, we argue that not only the context diversity but also the \emph{context informativeness} by language modeling, are both important factors to adapt CLIP to the zero-shot learning task. The insight behind this is that the sentence-level descriptions could contextualize compositional classes in a more fine-grained manner than the prior arts. Therefore, we propose to address the two factors by learning to \textbf{P}rompt the \textbf{L}anguage-\textbf{I}nformed \textbf{D}istributions ($\mathbb{PLID}$) for the CZSL task.

\textbf{Compositional Class Description.} To generate diverse and informative text descriptions for each compositional class, we adopt a similar way as~\cite{menon2022visual} by prompting an LLM that shows instruction-following capability. An example below shows the format of the LLM instruction. 
\begin{adjustwidth}{0cm}{0cm}
\begin{lstlisting}[breakatwhitespace=true]
Keywords: sliced, potato, picture
Output: The picture features a beautifully arranged plate of thinly sliced potatoes.
###
\end{lstlisting}
\end{adjustwidth}
See the Supplement~\ref{apd:llm_gen} for more details.
For each composition $y=(s,o)$, we generate $M$ descriptions denoted as $S^{(y)}=\{S^{(y)}_1,\ldots,S^{(y)}_M\}$ where $S^{(y)}_m$ is a linguistically complete sentence. Different to~\cite{menon2022visual} that aims to interpret the zero-shot recognition by attribute phrases from LLMs, we utilize the LLM-based sentence-level descriptions in the CZSL task for two benefits: (\textit{i}) provide diverse and informative textual context for modeling the class distributions, and (\textit{ii}) enhance the class embedding with fine-grained descriptive information. 

\textbf{Language-Informed Distribution (LID).} For both the image and text modalities, we use the frozen CLIP model and learnable feature enhancement modules to represent the visual and language features, which are also adopted in existing CZSL literature~\cite{dfsp_cvpr23,troika_arxiv23}. 

Specifically, for the text modality, each composition $y$ is tokenized and embedded by CLIP embedding layer and further prompted by concatenating with learnable context vectors, \ie, ``$[\mathbf{p}_1]\ldots[\mathbf{p}_L][\mathbf{s}][\mathbf{o}]$'', where $\mathbf{p}_{1:L}$ is initialized by ``\texttt{a photo of}'' and shared with all classes. Followed by the frozen CLIP text encoder $\mathcal{E}_T$, the embedding of class $y$ is  $\mathbf{q}_y = \mathcal{E}_T\left([\mathbf{p}_1]\ldots[\mathbf{p}_L][\mathbf{s}][\mathbf{o}]\right)$ where $\mathbf{q}_y\in\mathbb{R}^{d}$. Following the CZSL literature~\cite{xu2022prompting,dfsp_cvpr23}, here the soft prompt $\mathbf{p}_{1:L}$ and primitive embeddings $[\mathbf{s}][\mathbf{o}]$ are learnable while $\mathcal{E}_T$ is frozen in training. 

To simultaneously address the lack of diversity and informativeness of the soft prompts, we propose to formulate the class-specific distributions supported by the texts $S^{(y)}$ and learn to prompt these distributions. Specifically, we encode $S^{(y)}$ by the frozen CLIP text encoder: $\mathbf{D}^{(y)} = \mathcal{E}_T(S^{(y)})$, where $\mathbf{D}^{(y)}\in \mathbb{R}^{M\times d}$. Then, we use $\mathbf{D}^{(y)}$ to enhance $\mathbf{q}_y$ by $\mathbf{t}_y=\Psi_{\text{TFE}}(\mathbf{q}_y, \mathbf{D}^{(y)})$ where $\Psi_{\text{TFE}}$ is the text feature enhancement (\textbf{TFE}) implemented by a single-layer cross attention Transformer~\cite{vaswani2017attention}. 
Similarly, given an image $\mathbf{x}$, to mitigate the loss of fine-grained cues, we augment it with $N$ views to be $\mathbf{X}=\{\mathbf{x}^{(1)},\ldots,\mathbf{x}^{(N)}\}$. Followed by the frozen CLIP visual encoder $\mathcal{E}_V$, the feature of $\mathbf{x}$ is enhanced by $\mathbf{v}\!=\!\Psi_{\text{VFE}}(\mathcal{E}_V(\mathbf{x}),\mathcal{E}_V(\mathbf{X}))$ where $\Psi_{\text{VFE}}$ is the visual feature enhancement (\textbf{VFE}) by cross attention~\cite{vaswani2017attention}, implemented with the same structure as TFE for simplicity. 

\begin{wrapfigure}[12]{R}{0.4\linewidth}
\vspace{-5mm}
    \centering
    \includegraphics[width=\linewidth]{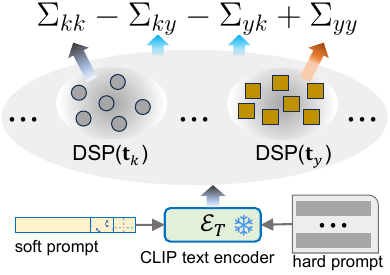}
    \caption{\small{Prompting for intra- and inter-class covariance optimization.}}
    \label{fig:prompt_dist}
\end{wrapfigure}

We treat the enhanced text feature $\mathbf{t}_y$ of class $y$ as the class mean and $\mathbf{t}_y+\mathbf{D}^{(y)}$ as the distribution support points (\textbf{DSP}) that follow the Gaussian $\mathcal{N}(\mathbf{t}_{y}, \boldsymbol{\Sigma}_{y})$ where $\boldsymbol{\Sigma}_{y}$ is the text variance of the class $y$. The motivation of $\mathbf{t}_y+\mathbf{D}^{(y)}$ is to enable the flexibility of DSP to traverse around in the $d$ dimensional space in training since $\mathbf{t}_y$ is trainable while $\mathbf{D}^{(y)}$ are pre-trained. For all $\lvert\mathcal{C}^{(s)}\rvert$ (denoted as $C$) seen compositional classes, we build joint Gaussian distributions $\mathcal{N}(\boldsymbol{\mu}_{1:C}, \boldsymbol{\Sigma}_{1:C})$ similar to ProDA~\cite{lu2022prompt}, where the means $\boldsymbol{\mu}_{1:C}\in\mathbb{R}^{C\times d}$ are given by $\mathbf{t}_y$ over $C$ classes, and the covariance $\boldsymbol{\Sigma}_{1:C}\in\mathbb{R}^{d\times C\times C}$ is defined across $C$ classes for each feature dimension from DSP. 

\textbf{Discussions.} Compared to the ProDA~\cite{lu2022prompt} that learns a collection of non-informative prompts, our DSPs are language-informed by $\mathbf{D}^{(y)}$ that provides more fine-grained descriptive information to help recognition and decomposition. Besides, our method is more parameter-efficient than ProDA since we only have a single soft prompt to learn. This is especially important for the CZSL task where there is a huge number of compositional classes. Lastly, we highlight the benefit of performing the intra- and inter-class covariance optimization induced by the learning objective of distribution modeling, which will be introduced below.

\textbf{Learning Objective.} Given the visual feature $\mathbf{v}\in\mathbb{R}^{d}$ of image $\textbf{x}$ and the text embeddings $\mathbf{t}_{1:C}$ from class-wise joint distributions $\mathcal{N}(\boldsymbol{\mu}_{1:C}, \boldsymbol{\Sigma}_{1:C})$, minimizing the cross-entropy loss is equivalent to minimizing the upper bound of negative log-likelihood (NLL):
\begin{equation}
    -\log\mathbb{E}_{\mathbf{t}_{1:C}} p(y|\mathbf{v}, \mathbf{t}_{1:C}) 
    \leq -\log \frac{\exp(h_y/\tau)}{\sum_{k=1}^{C} \exp((h_k + h_{k,y}^{(m)})/\tau)} \vcentcolon= \mathcal{L}_y(\mathbf{x},y),
\label{eq:loss}
\end{equation}
where the compositional logit $h_y=\texttt{cos}(\mathbf{v},\mathbf{t}_y)$, the pairwise margin $h_{k,y}^{(m)}=\mathbf{v}^\top\mathbf{A}_{k,y}\mathbf{v}/(2\tau)$ and $\mathbf{A}\in\mathbb{R}^{d\times C\times C}$ is given by $\mathbf{A}_{k,y}=\boldsymbol{\Sigma}_{kk}+\boldsymbol{\Sigma}_{yy}-\boldsymbol{\Sigma}_{ky}-\boldsymbol{\Sigma}_{yk}$. The covariance $\mathbf{A}_{k,y}$ indicates the correlation between the $k$-th out of $C$ classes and the target class $y$ on each of $d$ feature dimensions. The insight of minimizing $\mathcal{L}_y(\mathbf{x},y)$ is illustrated in \cref{fig:prompt_dist}, which encourages minimizing intra-class variance by $\boldsymbol{\Sigma}_{yy}$ and $\boldsymbol{\Sigma}_{kk}$, and maximizing inter-class separability indicated by $\boldsymbol{\Sigma}_{ky}$ and $\boldsymbol{\Sigma}_{yk}$. In Supplement~\ref{apd:cov_share}, we discuss the case when $C$ is too large to compute $\mathbf{A}$, our workaround by covariance sharing within each object group leads to negligible performance decrease.

\subsection{Primitives Decomposition for Fused Recognition}

\begin{wrapfigure}[10]{R}{0.4\linewidth}
\vspace{-5mm}
    \centering
    \includegraphics[width=\linewidth]{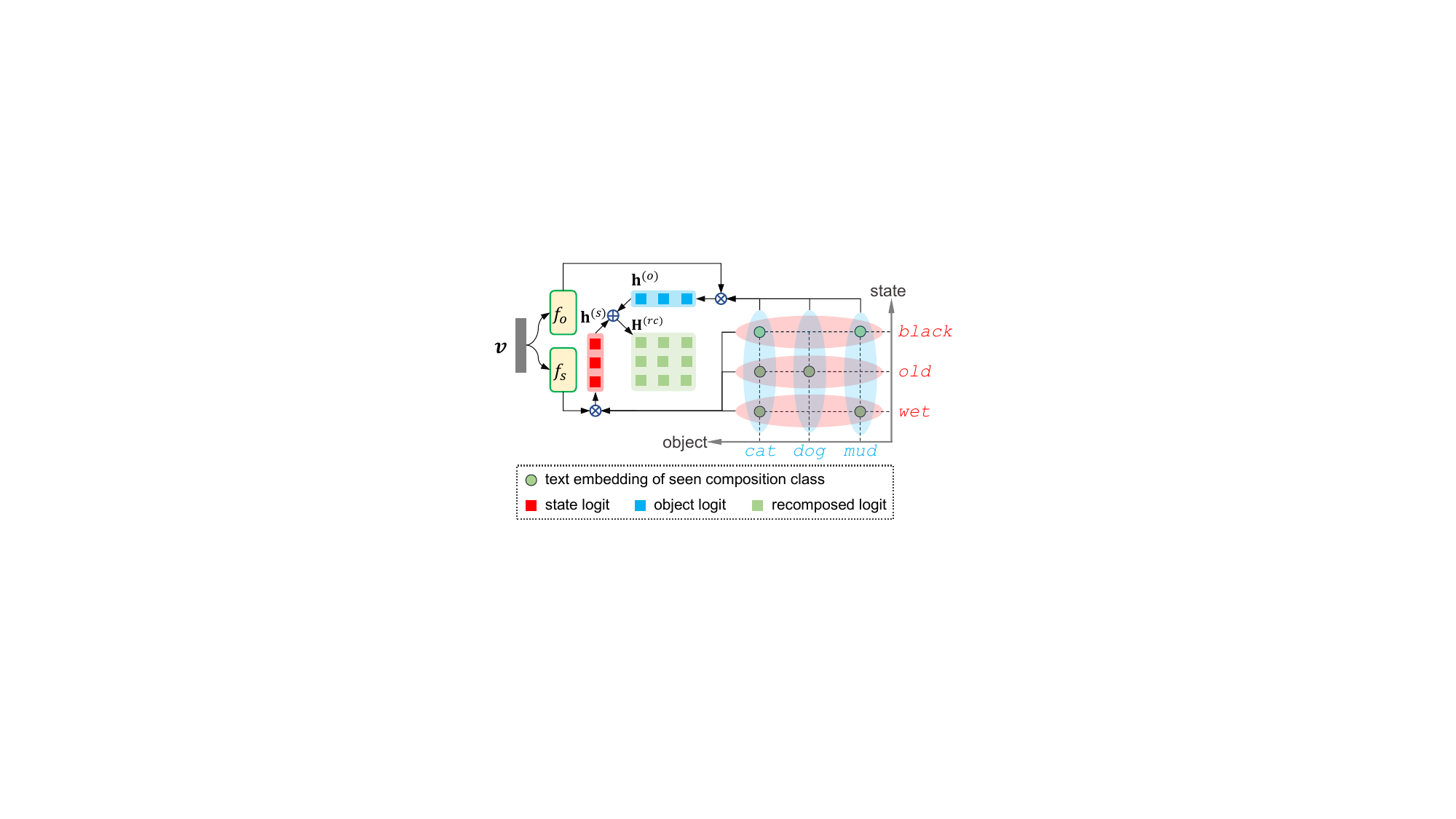}
    \captionsetup{font=small,aboveskip=-2pt}
    \caption{VLPD for recomposing.}
    \label{fig:vlpd}
\end{wrapfigure}

\textbf{Motivation.} Considering the fundamental challenge in the CZSL task, that the visual primitives are inherently entangled in an image, an unseen composition in testing can hardly be identified if its object (or its state) embedding is overfitted to the visual data of seen compositions. To this end, it is better to inherit the benefits of the decompose-recompose paradigm~\cite{zou2020compositional,karthik2022kg,liu2022simple} by decomposing visual features into simple primitives, \ie, states and objects, from which the recomposed decision can be leveraged for zero-shot recognition. Thanks to the compositionality of CLIP~\cite{wolff2023independent,trager2023linear}, such motivation can be achieved by the visual-language primitive decomposition (\textbf{VLPD}). See \cref{fig:vlpd} and we explain it below. Based on VLPD, we propose the stochastic logit mixup to fuse the directly learned compositions and the recomposed ones.

\textbf{VLPD.} Specifically, we use two parallel neural networks $f_s$ and $f_o$ to decompose $\mathbf{v}$ into the state visual feature $f_s(\mathbf{v})$ and object visual feature $f_o(\mathbf{v})$, respectively. To get the primitive-level supervisions, given the training compositions $\mathcal{C}^{(s)}$ (see the circle dots in \cref{fig:vlpd}), we group their enhanced embeddings $\{\mathbf{t}_y\}$ over the subset $\mathcal{Y}_o$, in which all compositions share the same given object $o$ (see vertical ellipses in \cref{fig:vlpd}), and group $\{\mathbf{t}_y\}$ over the subset $\mathcal{Y}_s$, in which all compositions share the same given state $s$ (see horizontal ellipses in \cref{fig:vlpd}). Thus, given a state $s$ and an object $o$, the predicted object logit $h_s$ and state logit $h_o$ are computed by 
\begin{equation}
    h_s = \texttt{cos}\left(f_s(\mathbf{v}),\frac{1}{\lvert\mathcal{Y}_s\rvert} \sum_{y\in \mathcal{Y}_s} \mathbf{t}_y\right), \;\;\;\;\; h_o = \texttt{cos}\left(f_o(\mathbf{v}),\frac{1}{\lvert\mathcal{Y}_o\rvert} \sum_{y\in \mathcal{Y}_o} \mathbf{t}_y\right).
\label{eq:hsho}
\end{equation}
Different from DFSP~\cite{dfsp_cvpr23} that only decomposes text features, we additionally use $f_s$ and $f_o$ to decompose visual features $\mathbf{v}$ and empirically show the superiority of performing both visual and language decomposition (see~\cref{tab:vlpd}).

Following the spirit of distribution modeling, we also introduce the distributions over state and object categories, where the corresponding DSP, denoted as $\mathbf{D}^{(s)}$ and $\mathbf{D}^{(o)}$, are obtained by grouping $\mathbf{D}^{(y)}$ over $\mathcal{Y}_s$ and $\mathcal{Y}_o$, respectively. This leads to the following upper-bounded cross-entropy losses:
\begin{equation}
\begin{split}
    & \mathcal{L}_s(x,s) = -\log \frac{\exp(h_s/\tau)}{\sum_{k=1}^{\lvert\mathcal{S}\rvert} \exp((h_k + h_{k,s}^{(m)})/\tau)}, \\
    & \mathcal{L}_o(x,o) = -\log \frac{\exp(h_o/\tau)}{\sum_{k=1}^{\lvert\mathcal{O}\rvert} \exp((h_k + h_{k,o}^{(m)})/\tau)},
\end{split}
\label{eq:loss_so}
\end{equation}
where $h_{k,s}^{(m)}$ and $h_{k,o}^{(m)}$ are determined the same way as $h_{k,y}^{(m)}$ in Eq.~(\ref{eq:loss}). See details in Supplement~\ref{sec:gauss_primitive}. In this way, the merits of language-informed distribution modeling, \ie, the inter- and intra-class covariance optimization constraints, can be introduced into primitive space for fused recognition as introduced below. 

\textbf{Composition Fusion.} With the individually supervised $f_s$ and $f_o$, we have $p(y|\mathbf{v}) = p(s|\mathbf{v})\cdot p(o|\mathbf{v})$ according to conditional independence, that induces $p(y|\mathbf{v}) \propto \exp((h_s + h_o)/\tau)$. Therefore, the recomposed logit matrix $\mathbf{H}^{(rc)}\in\mathbb{R}^{\lvert\mathcal{S}\rvert\times \lvert\mathcal{O}\rvert}$ is a Cartesian (element-wise combinatorial) sum between $\mathbf{h}^{(s)}\in\mathbb{R}^{\lvert\mathcal{S}\rvert}$ and $\mathbf{h}^{(o)}\in\mathbb{R}^{\lvert\mathcal{O}\rvert}$, \ie, $\mathbf{H}^{(rc)} = \mathbf{h}^{(s)} \oplus \mathbf{h}^{(o)\top}$, where $\mathbf{h}^{(s)}$ contains all state logits and $\mathbf{h}^{(o)}$ contains all object logits. See the red and blue squares in \cref{fig:vlpd}.

Given the recomposed logit $h_y^{(rc)}\in \mathbf{H}^{(rc)}$ and the directly learned compositional logit $h_y$ by~\cref{eq:loss}, we propose to stochastic fusion method in training by sampling a coefficient $\lambda$ from a Beta prior distribution:
\begin{equation}
    \tilde{h}_y = (1-\lambda) h_y + \lambda h_y^{(rc)}, \;\;\; \lambda \sim \text{Beta}(a,b),
\end{equation}
where $(a,b)$ are hyperparameters indicating the prior preference for each decision. In training, we replace the $h_y$ and $h_k$ of Eq.~(\ref{eq:loss}) with the mixed logit $\tilde{h}_y$ and $\tilde{h}_k$, respectively. In testing, no stochasticity is needed so we use the Beta expectation of $\lambda$ which is $a/(a+b)$ to get the fused logit $\tilde{h}_y$.

The insights behind the stochasticity are that the Beta distribution indicates a prior preference to $h_y$ or $h_y^{(rc)}$. It provides the flexibility of which compositional decision to trust in, and the stochasticity of the coefficient $\lambda$ inherently introduces a regularization effect in training ~\cite{mixup_jmlr22}. Moreover, compared to softmax probability mixup~\cite{troika_arxiv23}, our logit mixup avoids the limitation of softmax normalization over a huge number of compositional classes, that rich information of class relationship is lost after softmax normalization according to \cite{bang2022logit}. Such class relationships are even more important in the CZSL problem as indicated in \cite{naeem2021learning}. 

\section{Experiments}

\newcolumntype{g}{>{\columncolor{Gray}}c}
\begin{table}[t]
\captionsetup{font=footnotesize,skip=0pt}
\caption{CZSL results of Closed- and Open-World settings on three datasets. Baseline results are from published literature except for ProDA. Note that ``--'' indicates no results reported by the PCVL paper or not applicable by ProDA for more than 278K compositional classes on the C-GQA dataset.}
\label{tab:main}
\centering
\small
\setlength{\tabcolsep}{0.28mm}
\setlength{\extrarowheight}{0.1mm}
\begin{tabular}{cllcccclcccclcccc}
  \toprule
  \multicolumn{2}{c}{\multirow{2}{*}{Method}} && \multicolumn{4}{c}{\textit{MIT-States}} &$\;\;$& \multicolumn{4}{c}{\textit{UT-Zappos}} &$\;\;$& \multicolumn{4}{c}{\textit{C-GQA}}\\ 
  \cmidrule{4-7} \cmidrule{9-12} \cmidrule{14-17}
  &  && S & U & H & AUC && S & U & H & AUC && S & U & H & AUC \\ 
  \midrule
\multirow{8}{*}{\parbox{1cm}{Closed}} 
 & CLIP~\cite{CLIP} && 30.2  & 46.0 & 26.1 & 11.0 && 15.8  & 49.1 & 15.6 & 5.0 && 7.5  & 25.0 & 8.6 & 1.4  \\
 & CoOp~\cite{zhou2022coop} &&34.4 & 47.6 &29.8 &13.5  &&52.1 &49.3 &34.6 &18.8  &&20.5 & {26.8}  &17.1 &4.4  \\
 & ProDA\footnotemark[1]~\cite{lu2022prompt} && 37.4 & 51.7 & 32.7 & 16.1 &&63.7 & 60.7 & 47.6 & 32.7 && -- & -- & -- & -- \\
 & CSP~\cite{csp_iclr23} &&  {46.6}   &  {49.9}   &  {36.3}  &  {19.4} 
  &&  {64.2}  &  {66.2}  &  {46.6} &  {33.0} 
  &&  {28.8}   &  {26.8}   &  {20.5}  &  {6.2}  \\
 & PCVL~\cite{xu2022prompting} && 48.5 & 47.2 & 35.3 & 18.3 && 64.4 & 64.0 & 46.1 & 32.2 && -- & -- & -- & -- \\
 & HPL~\cite{wang2023hpl} && 47.5 & 50.6 & 37.3 & 20.2  && 63.0 & 68.8 & 48.2 & 35.0 && 30.8 & 28.4 & 22.4 & 7.2 \\
 & DFSP~\cite{dfsp_cvpr23} && 46.9 & 52.0 & 37.3 & 20.6 && 66.7 & \textbf{71.7} & 47.2 & 36.0 && 38.2 & 32.0 & 27.1 & 10.5\\
 &\cellcolor{Gray}\textbf{$\mathbb{PLID}$} 
 &\cellcolor{Gray}&\cellcolor{Gray}\textbf{49.7} &\cellcolor{Gray}\textbf{52.4} &\cellcolor{Gray}\textbf{39.0} &\cellcolor{Gray}\textbf{22.1}
 &\cellcolor{Gray}&\cellcolor{Gray}\textbf{67.3} &\cellcolor{Gray}68.8 &\cellcolor{Gray}\textbf{52.4} &\cellcolor{Gray}\textbf{38.7}
 &\cellcolor{Gray}&\cellcolor{Gray}\textbf{38.8} &\cellcolor{Gray}\textbf{33.0} &\cellcolor{Gray}\textbf{27.9} &\cellcolor{Gray}\textbf{11.0} \\ \hline
\multirow{8}{*}{\parbox{1cm}{Open}} 
& CLIP~\cite{CLIP} && 30.1 & 14.3 & 12.8 & 3.0 && 15.7 & 20.6 & 11.2 & 2.2 && 7.5 & 4.6 & 4.0 & 0.3\\
 & CoOp~\cite{zhou2022coop} &&34.6 &9.3 &12.3 &2.8 &&52.1 &31.5  &28.9 &13.2  &&21.0  &4.6  &5.5  &0.7  \\
 & ProDA\footnotemark[1]~\cite{lu2022prompt} && 37.5 & 18.3 & 17.3 & 5.1 &&63.9 & 34.6 & 34.3 & 18.4 && -- & -- & -- & -- \\
 & CSP~\cite{csp_iclr23} &&  {46.3}  &  {15.7}    &  {17.4}  &  {5.7}  &&  {64.1}  &  {44.1}   &  {38.9}  &  {22.7}  &&  {28.7}  &  {5.2}   &  {6.9}   &  {1.2}  \\
 & PCVL~\cite{xu2022prompting} && 48.5 & 16.0 & 17.7 & 6.1 && 64.6 & 44.0 & 37.1 & 21.6 && -- & -- & -- & -- \\
 & HPL~\cite{wang2023hpl} && 46.4 & \textbf{18.9} & 19.8 & 6.9 && 63.4 & 48.1 & 40.2 & 24.6 && 30.1 & 5.8 & 7.5 & 1.4 \\
 & DFSP~\cite{dfsp_cvpr23} && 47.5 & 18.5 & 19.3 & 6.8 && 66.8 & \textbf{60.0} & 44.0 & 30.3 && 38.3 & 7.2 & 10.4 & 2.4 \\
 &\cellcolor{Gray}\textbf{$\mathbb{PLID}$} 
 &\cellcolor{Gray}&\cellcolor{Gray}\textbf{49.1} &\cellcolor{Gray}{18.7} &\cellcolor{Gray}\textbf{20.4} &\cellcolor{Gray}\textbf{7.3} 
 &\cellcolor{Gray}&\cellcolor{Gray}\textbf{67.6} &\cellcolor{Gray}55.5 &\cellcolor{Gray}\textbf{46.6} &\cellcolor{Gray}\textbf{30.8}
 &\cellcolor{Gray}&\cellcolor{Gray}\textbf{39.1} &\cellcolor{Gray}\textbf{7.5} &\cellcolor{Gray}\textbf{10.6} &\cellcolor{Gray}\textbf{2.5} \\ \hline
\end{tabular}
\end{table}

\textbf{Datasets.} We perform experiments on three CZSL datasets, \ie, MIT-States~\cite{mitstates}, UT-Zappos~\cite{utzappos}, and C-GQA~\cite{naeem2021learning}, following the standard splitting protocols in CZSL literature~\cite{purushwalkam2019task,csp_iclr23,dfsp_cvpr23}. MIT-States consists of 115 states and 245 objects, with 53,753 images in total. Following~\cite{purushwalkam2019task,csp_iclr23,dfsp_cvpr23}, it is split into 1,262 seen and 300/400 unseen compositions for training and validation/testing, respectively. UT-Zappos contains 16 states and 12 objects for 50,025 images in total, and it is split into 83 seen and 15/18 unseen compositions for training and validation/testing. C-GQA contains 453 states and 870 objects for 39,298 images, and it is split into 5,592 seen and 1,040/923 unseen compositions for training and validation/testing, respectively, resulting in 7,555 and 278,362 target compositions in closed- and open-world settings.

\textbf{Evaluation.} 
We report the metrics in both closed-world (\textbf{CW}) and open-world (\textbf{OW}) settings, including the best seen accuracy (\textbf{S}), the best unseen accuracy (\textbf{U}), the best harmonic mean (\textbf{H}) between the seen and unseen accuracy, and the area under the curve (\textbf{AUC}) of unseen versus seen accuracy. For OW evaluation, following the CSP~\cite{csp_iclr23}, we adopt the feasibility calibration by GloVe~\cite{pennington2014glove} to filter out infeasible compositions.

\textbf{Implementation Details.} We implement the $\mathbb{PLID}$ based on the CSP codebase in PyTorch. The CLIP architecture ViT-L/14 is used by default. On the MIT-States, we generate $M=64$ texts and augment an image with $N=8$ views, and adopt $\text{Beta}(1,9)$ as prior. The dropout rates of cross-attention layers in TFE and VFE are set to 0.5, and the dropout rate to 0.3 for the learnable state and object embeddings. For the soft prompt embeddings, we set the context length of the text encoder to 8 for all datasets. Following~\cite{dfsp_cvpr23}, we use Adam optimizer with base learning rate 5e-5 and weight decay 2e-5, and step-wise decay it with the factor of 0.5 every 5 training epochs for a total of 20 epochs. Complete training hyperparameters on three datasets are in the Supplement~\ref{apd:impl}. 

\footnotetext[1]{\scriptsize{ProDA is re-implemented for the CZSL setting. Limited by the GPU memory, ProDA is not applicable to the C-GQA dataset which consists of more than 278K compositional classes.}}

\subsection{Main Results}

The results are reported in~\cref{tab:main}. We compare with the CZSL baselines that are developed on the same frozen CLIP model. The table shows that under both the closed-world and open-world test settings, our proposed $\mathbb{PLID}$ method achieves the best performance in most metrics on the three datasets. Note that ProDA~\cite{lu2022prompt} also formulates the class-wise Gaussian distributions to address the intra-class diversity, but it can only outperform CLIP and CoOp on all metrics. This indicates the importance of both diversity and informativeness for the CZSL task. On the UT-Zappos dataset, the $\mathbb{PLID}$ outperforms the DFSP in terms of S, H, and AUC by 0.6\%, 5.2\%, and 2.7\% respectively, while inferior to the DFSP on the best unseen metric. The potential reason is that DFSP fuses the text features into the image images, which better preserves the generalizability of CLIP for the small downstream UT-Zappos dataset. Note that the HPL method uses prompt learning and recognition at both compositional and primitive levels, but it performs only slightly better than CSP and way worse than our method, indicating that traditional prompt learning helps but is not enough to adapt the CLIP model to the CZSL task.

\subsection{Model Analysis}

\begin{table}[t]
\small
\setlength{\tabcolsep}{1.7mm}
\setlength{\extrarowheight}{0.1mm}
    \centering
    \begin{tabular}{c|cccc|cccc} 
        \toprule
         & LID & FE  & OPT  & PDF & $\text{H}_\text{cw}$ & $\text{AUC}_\text{cw}$  & $\text{H}_\text{ow}$ & $\text{AUC}_\text{ow}$ \\
         \hline
        (a) &   &   &   &     &  35.41 & 18.56   & 17.37  & 5.56  \\
        (b) &  \cmark &     &   &     &  37.06	& 20.43 & 18.65 & 6.50  \\
        (c) &  \cmark & \cmark  &   &     &  37.87 & 21.09 & 19.70 & 6.95 \\
        (d) &  \cmark & \cmark  & \cmark  &     &  38.80 & 21.67 & 19.61 & 7.01 \\
        (e) &  \cmark & \cmark  & \cmark  & \cmark  &  \textbf{38.97} & \textbf{22.12} & \textbf{20.41} & \textbf{7.34}  \\
         \bottomrule
    \end{tabular}
    \captionsetup{font=small,aboveskip=-5pt}
    \caption{\textbf{Ablation study}. (a): the baseline that uses mean pooling of text embeddings from T5-generated sentences. (b): add language-informed distribution (LID). (c): use text and visual feature enhancement module (FE). (d): change the LLM from T5-base to the OPT-1.3B. (e): apply primitive decomposition for fused decision (PDF).}
    \label{tab:ablation}
\end{table}

To comprehensively analyze the proposed $\mathbb{PLID}$, we perform extensive ablation study and design analysis on the middle-sized MIT-States dataset in this section. More ablation results are provided in the Supplement~\ref{apd:impl}.

\textbf{Major Components.} In~\cref{tab:ablation}, we show the contribution of the major components in  the $\mathbb{PLID}$ model. It is clear that they are all beneficial. We highlight some important observations: (1) The LID method in row (b) significantly improves the performance compared to the baseline (a) that does not formulate Gaussian distribution in training, and they are much better than ProDA (20.43\% vs 16.1\% of $\text{AUC}_{\text{cw}}$) when referring to~\cref{tab:main}. This implies that addressing the context \textbf{diversity} by modeling the Gaussian distribution like the ProDA is not sufficient, but context \textbf{informativeness} is critical and preferred for the CZSL task. (2) Rows (c)(d) show that feature enhancement (FE) and the better LLM OPT-1.3B can also bring performance gains. (3) Rows (e) show that the primitive decomposition for fused decision (PDF) could further improve the CZSL performance in both closed- and open-world settings. In the following paragraphs, we further validate the effect or design choices of these components in detail.

\begin{table}[t]
\CenterFloatBoxes
\begin{floatrow}
\capbtabbox{%
    \centering
    \footnotesize
    \setlength{\tabcolsep}{0.1mm}
    \setlength{\extrarowheight}{0.1mm}
    \begin{tabular}{ccllcclcc}
      \toprule
      $\mathcal{N}_s$ & $\mathcal{N}_o$ &$\mathcal{N}_y$  
       && $\text{H}_\text{cw}$ & $\text{AUC}_\text{cw}$  && $\text{H}_\text{ow}$ & $\text{AUC}_\text{ow}$ \\ 
      \midrule
       &  &  && 38.44 & 21.67 && 19.53 & 6.99 \\
      \cmark &\cmark &  && 38.30 & 21.62 && 19.49 & 6.95 \\
        &  &\cmark && 38.49 & 21.90 && 19.93 & 7.20 \\
       \cellcolor{Gray}\cmark &\cellcolor{Gray}\cmark &\cellcolor{Gray}\cmark &\cellcolor{Gray}&\cellcolor{Gray}{\textbf{38.97}} &\cellcolor{Gray}{\textbf{22.12}} &\cellcolor{Gray}&\cellcolor{Gray}\textbf{20.41}&\cellcolor{Gray}\textbf{7.34} \\
      \bottomrule
    \end{tabular}
}{\caption{\footnotesize Effect of LID on states ($\mathcal{N}_s$), objects ($\mathcal{N}_o$), and compositions ($\mathcal{N}_y$). The first row indicates the model without LID.}
\label{tab:lids}
}
\hfill
\capbtabbox{%
    \centering
    \footnotesize
    \setlength{\tabcolsep}{0.2mm}
    \setlength{\extrarowheight}{0.1mm}
    \begin{tabular}{llcccc}
      \toprule
      LLMs && $\text{H}_\text{cw}$ & $\text{AUC}_\text{cw}$  & $\text{H}_\text{ow}$ & $\text{AUC}_\text{ow}$ \\ 
      \midrule
      Mistral-7B &&  37.22  & 20.78   & 19.22   & 6.74  \\%
     GPT-3.5 && 37.38   & 20.61   & 19.38  & 6.80   \\
     T5-base && 38.41 & 21.53 & \textbf{20.46} & {7.34} \\
     \cellcolor{Gray}OPT-1.3B &\cellcolor{Gray}&\cellcolor{Gray}\textbf{38.97} &\cellcolor{Gray}\textbf{22.12} &\cellcolor{Gray}{20.41} &\cellcolor{Gray}\textbf{7.34} \\ 
     \bottomrule
    \end{tabular}
}{\caption{\footnotesize Effect of LLMs. Note that GPT-3.5 is not open-sourced so that we use its API call to get text descriptions.}
\label{tab:llm}
}
\end{floatrow}
\end{table}

\textbf{Effect of LID.} In~\cref{tab:lids}, we investigate at which semantic level the language-informed distribution (LID) should be applied. Denote the Gaussian distribution on state, object, and composition as $\mathcal{N}_s$, $\mathcal{N}_o$, and $\mathcal{N}_y$, respectively. The~\cref{tab:lids} results clearly show the superiority of applying LID on all three semantic levels. This indicates the generality of LID towards many potential zero-shot or open-vocabulary recognition problems.

\textbf{Effect of LLM.} In~\cref{tab:llm}, we analyze the choice of LLMs by comparing $\mathbb{PLID}$ variants using different LLMs, including the T5-base~\cite{T5_JMLR20}, OPT-1.3B~\cite{OPT_arXiv22}, GPT-3.5~\cite{gpt35}, and Mistral-7B~\cite{mistral7b}. It shows the performance varies across different LLMs. Note that the capacity of GPT-3.5 and Mistral-7B on general language processing tasks is much better than T5-base and OPT-1.3B. However, we do not see improvements by using these generally larger and better LLMs, but a small OPT-1.3B is sufficient to achieve the best performance. We provide some examples of the generated texts by these LLMs in Supplement~\ref{apd:llm_gen}.

\begin{table}[t]
\BottomFloatBoxes
\begin{floatrow}
\resizebox{0.5\linewidth}{!}{
\capbtabbox{%
    \centering
    \scriptsize
    \setlength{\tabcolsep}{0.1mm}
    \setlength{\extrarowheight}{0.1mm}
    \begin{tabular}{cc|c|cccc}
      \toprule
       \text{TFE} & \text{VFE}  & layers & $\text{H}_\text{cw}$ & $\text{AUC}_\text{cw}$  & $\text{H}_\text{ow}$ & $\text{AUC}_\text{ow}$ \\ 
      \midrule
       \cmark &  & 1  & 37.89 & 21.07  & 19.37  & 6.78  \\
       & \cmark  & 1 & 37.48 & 21.04 & 19.43 & 6.72 \\
       \cmark & \cmark & 3  & 37.46	& 20.65 & 19.15	& 6.70 \\
       \cellcolor{Gray}\cmark &\cellcolor{Gray}\cmark  &\cellcolor{Gray}1 &\cellcolor{Gray}{\textbf{38.97}} &\cellcolor{Gray}{\textbf{22.12}} &\cellcolor{Gray}\textbf{20.41}&\cellcolor{Gray}\textbf{7.34} \\
      \bottomrule
    \end{tabular}
}{\caption{\footnotesize Design choices of feature enhancement (FE). We explore the use of text or visual feature enhancement (TFE/VFE) and the number of their cross-attention layers.}
\label{tab:fe}
}
}
\hfill
\resizebox{0.51\linewidth}{!}{
\capbtabbox{%
    \centering
    \scriptsize
    \setlength{\tabcolsep}{0.4mm}
    \setlength{\extrarowheight}{0.1mm}
    \begin{tabular}{cc|cc|cccc}
      \toprule
      \multicolumn{2}{c|}{VLPD} & \multicolumn{2}{c|}{fusion} & \multirow{2}{*}[-0.5ex]{$\text{H}_\text{cw}$} & \multirow{2}{*}[-0.5ex]{$\text{AUC}_\text{cw}$}  & \multirow{2}{*}[-0.5ex]{$\text{H}_\text{ow}$} & \multirow{2}{*}[-0.5ex]{$\text{AUC}_\text{ow}$} \\ 
      \cline{1-4}
       \text{text} & \text{image} & det. & stoc. & \\
      \midrule
       &  &  & \cmark & 37.94  & 20.98  & 19.67  & 6.98  \\
       \cmark &  &  & \cmark & 38.40  &  21.31 &  19.99 &  7.13 \\
       \hline
       \cmark & \cmark &  &  & 38.42 & 21.69 & 20.24 & 7.31 \\
       \cmark & \cmark & \cmark  &  & 38.67 & 21.90 & 19.99 & 7.15  \\
       \hline
       \cellcolor{Gray}\cmark &\cellcolor{Gray}\cmark  &\cellcolor{Gray} &\cellcolor{Gray}\cmark &\cellcolor{Gray}{\textbf{38.97}} &\cellcolor{Gray}{\textbf{22.12}} &\cellcolor{Gray}\textbf{20.41}&\cellcolor{Gray}\textbf{7.34} \\
      \bottomrule
    \end{tabular}
}{\caption{\footnotesize Effect of VLPD and fusion strategies. We explore the modalities (text or image) of the decomposition, and whether deterministic (det.) or stochastic (stoc.) compositional fusion. 
    }
\label{tab:vlpd}
}
}
\end{floatrow}
\end{table}
\textbf{TFE and VFE.} In~\cref{tab:fe}, we explore the design choices of the text and visual feature enhancement (TFE and VFE) modules. The results show that using one layer of randomly initialized cross-attention for both TFE and VFE performs the best. 
Using more cross-attention layers will cause a significant performance drop (see the 3rd row). We attribute the cause to the over-fitting issue when more learnable parameters are introduced to aggregate text descriptions or visual features.

\textbf{VLPD and Fusion.} In~\cref{tab:vlpd}, we validate the design choices of visual language primitive decomposition (VLPD) and the stochastic compositional fusion. Compared with the results of the first two rows, it shows clear advantages of primitive decomposition over both image and text modalities. Note that DFSP~\cite{dfsp_cvpr23} also has primitive decomposition but only on text modality. Our better performance than DFSP and the results in~\cref{tab:vlpd} thus tell the need for decomposition on both visual and image. Besides, to validate our stochastic compositional fusion, we compare it with the model without fusion in the 3rd row and the model with only deterministic fusion (weighted average without Beta sampling) in the 4th row. They also show the benefit of fusion with stochasticity.

\begin{figure}[t]
\begin{minipage}{0.5\textwidth}
        \centering
        \subcaptionbox{\vspace{2mm}AUC vs. $M$\label{fig:auc_m}}{
            \includegraphics[width=0.45\linewidth]{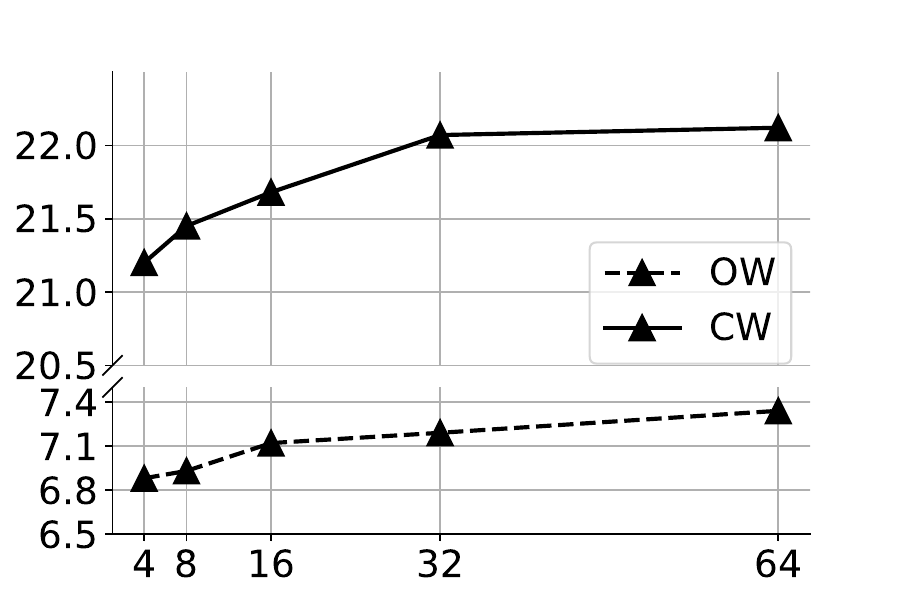}
        }
        \subcaptionbox{\vspace{2mm}AUC vs. $N$\label{fig:auc_n}}{
            \includegraphics[width=0.45\linewidth]{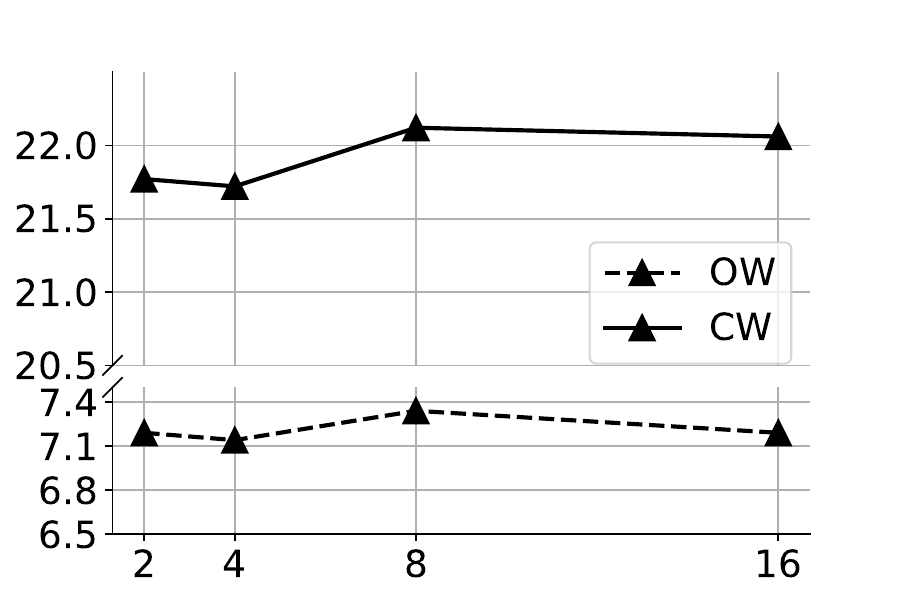}
        }
        \captionsetup{font=footnotesize,aboveskip=2pt}
        \RawCaption{\caption{Impact of $M$ and $N$. We set $N=8$ for the Fig.~\ref{fig:auc_m}, while we set $M=64$ for the Fig.~\ref{fig:auc_n}.}
        \label{fig:impact_mn}}
\end{minipage}
\hfill
\begin{minipage}{0.45\textwidth}
        \centering
        \subcaptionbox{\vspace{4mm}best HM\label{fig:ab_hm}}{
            \includegraphics[width=0.45\linewidth]{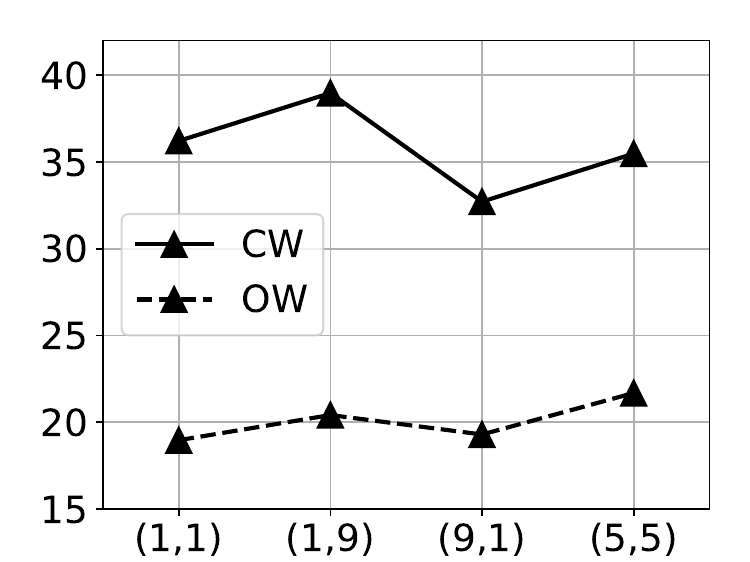}
        }
        \subcaptionbox{\vspace{4mm}best AUC\label{fig:ab_auc}}{
            \includegraphics[width=0.45\linewidth]{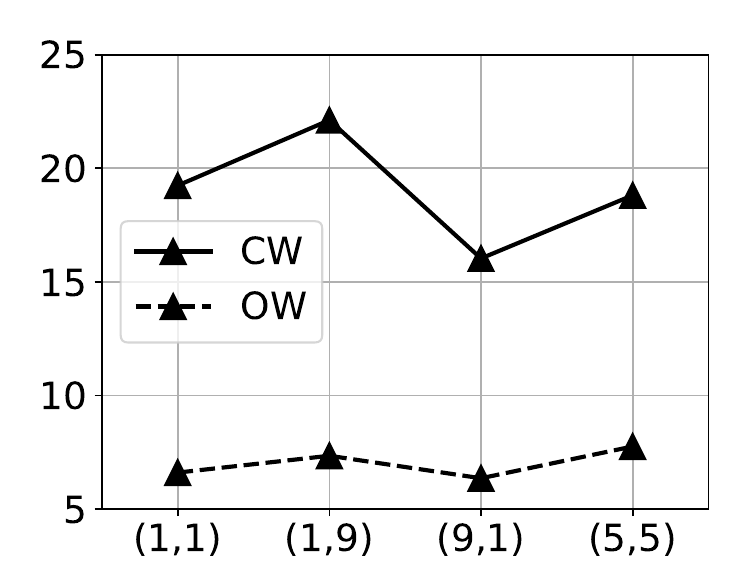}
        }
        \captionsetup{font=footnotesize,aboveskip=-6pt}
        \RawCaption{\caption{Impact of $(a,b)$. Here $(1,1)$ implies random sampling while $(5,5)$ implies equally trusted.}
        \label{fig:impact_ab}}
\end{minipage}
\end{figure}
\begin{figure}[t]
    \centering
    \begin{minipage}{0.32\linewidth}
        \subcaptionbox{LLM \textbf{compositions}\label{fig:tsne_txt}}{
            \includegraphics[width=\linewidth]{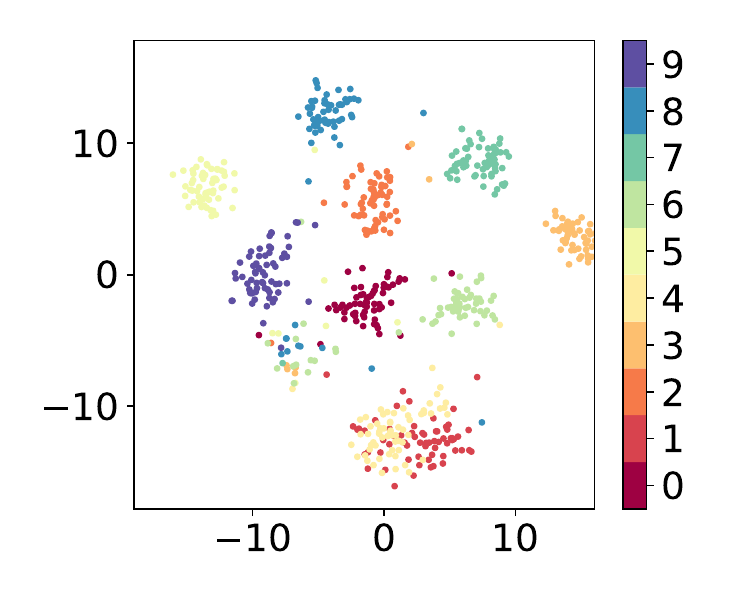}
        }
    \end{minipage}
    \hfill
    \begin{minipage}{0.32\linewidth}
        \subcaptionbox{LLM \textcolor{red}{\textbf{states}}\label{fig:tsne_txt_state}}{
            \includegraphics[width=\linewidth]{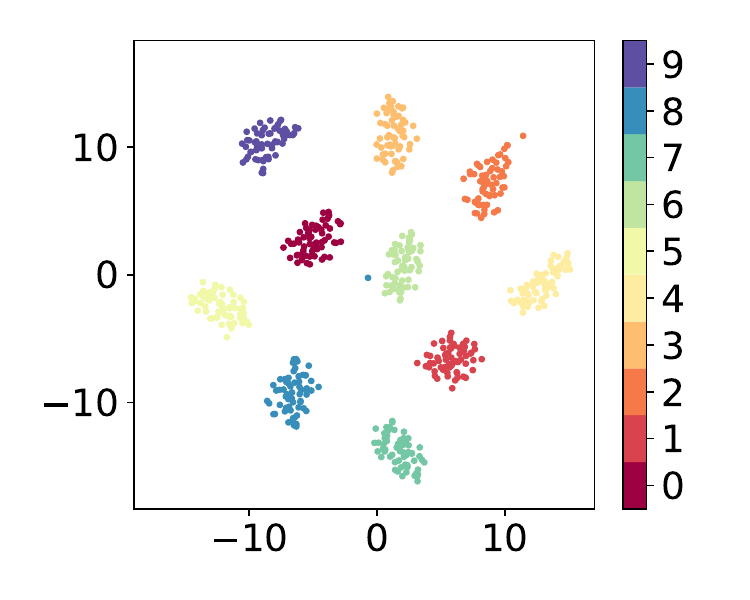}
        }
    \end{minipage}
    \hfill
    \begin{minipage}{0.32\linewidth}
        \subcaptionbox{LLM \textcolor{blue}{\textbf{objects}}\label{fig:tsne_txt_obj}}{
            \includegraphics[width=\linewidth]{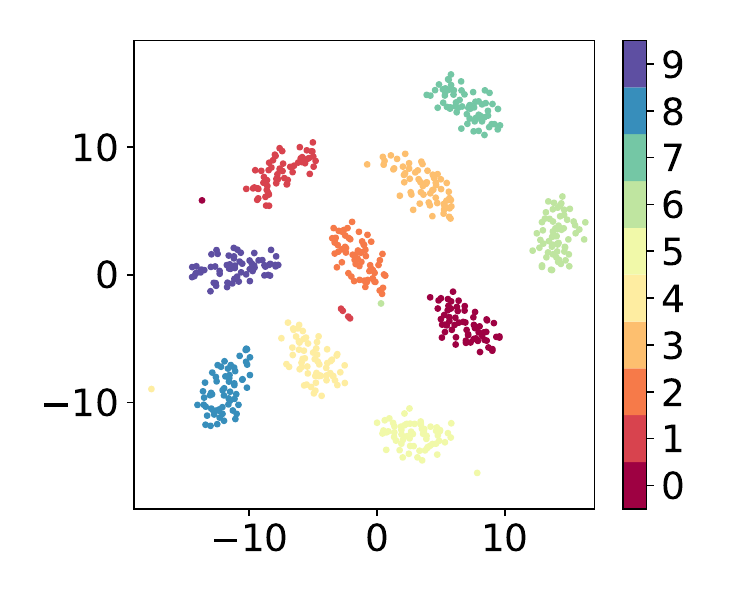}
        }
    \end{minipage}
    \vspace*{-2mm}
    \begin{minipage}{0.32\linewidth}
        \subcaptionbox{Learned \textbf{composition} DSP\label{fig:tsne_dsp}}{
            \includegraphics[width=\linewidth]{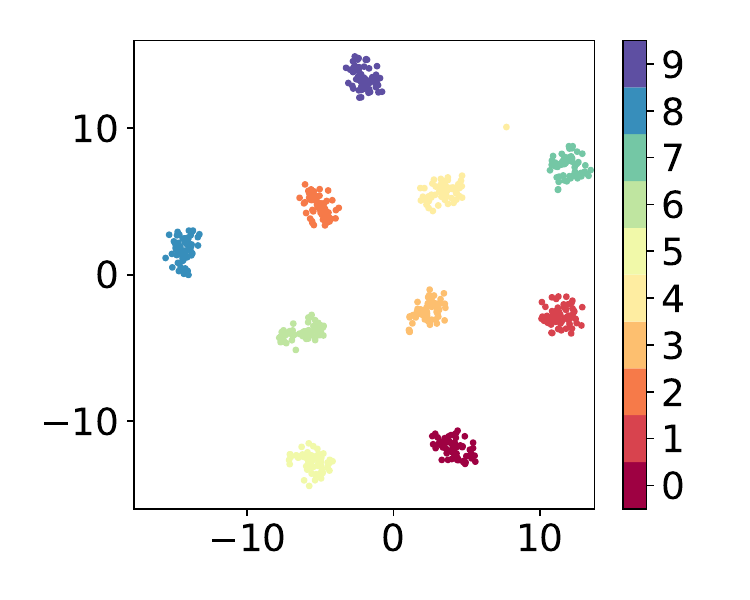}
        }
    \end{minipage}
    \hfill
    \begin{minipage}{0.32\linewidth}
        \subcaptionbox{Learned \textcolor{red}{\textbf{state}} DSP\label{fig:tsne_dsp_state}}{
            \includegraphics[width=\linewidth]{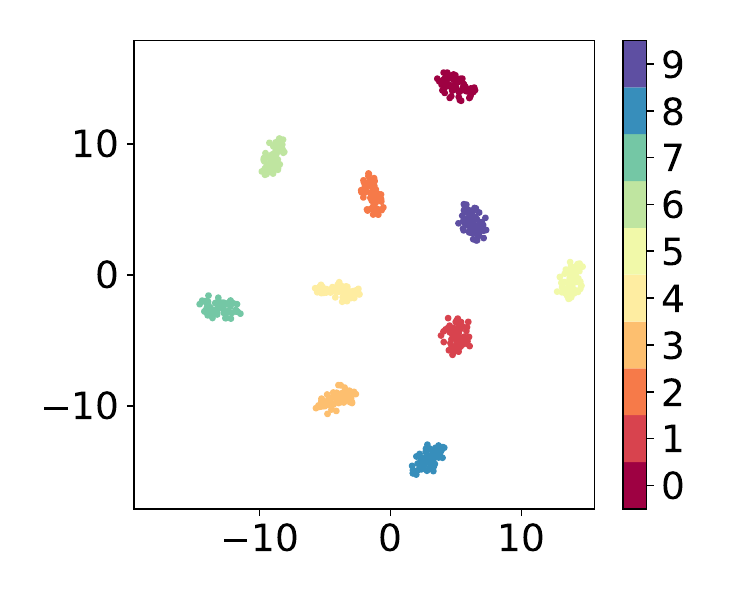}
        }
    \end{minipage}
    \hfill
    \begin{minipage}{0.32\linewidth}
        \subcaptionbox{Learned \textcolor{blue}{\textbf{object}} DSP\label{fig:tsne_dsp_obj}}{
            \includegraphics[width=\linewidth]{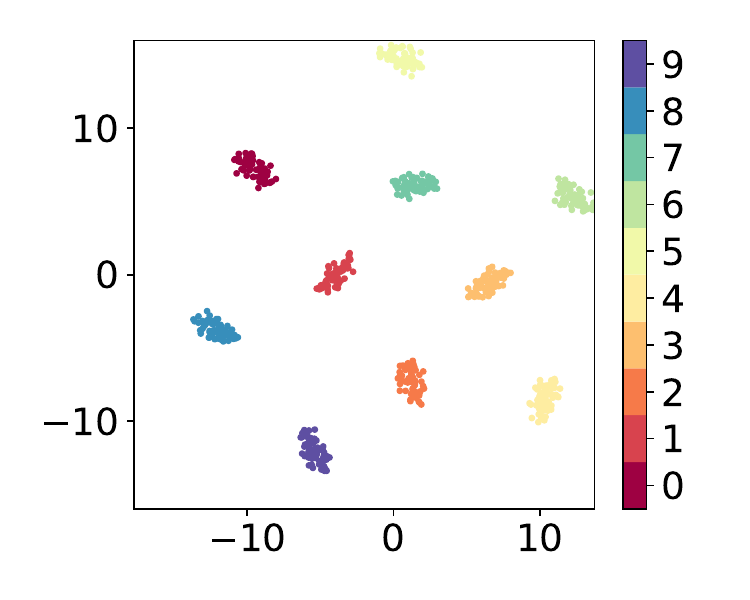}
        }
    \end{minipage}
    \caption{tSNE visualization of the text embeddings with (the 2nd row) and without (the 1st row) learnable distribution modeling over compositions (the 1st column), states (the 2nd column), and objects (the 3rd column). This figure clearly shows that our method achieves good performance by distribution modeling.}
    \label{fig:tsne_all}
\end{figure}

\textbf{Hyperparameters.} In \cref{fig:impact_mn}, we show the impact of the number of generated text descriptions $M$ and the number of augmented image views $N$. It shows that the best performance is achieved when $M=64$ and $N=8$. We note that more augmented image views slightly decrease the performance, which could be attributed to the overfitting of the seen compositions. 
In \cref{fig:impact_ab}, we show the impact of the Beta prior parameters $(a,b)$. We set them to $(1,1)$ for random sampling, $(1,9)$ for preference to the composition, $(9,1)$ for preference to re-composition, and $(5,5)$ for equal preference, respectively. It reveals that trusting more of the directly learned composition by $\text{Beta}(1,9)$ achieves the best results.

\begin{figure}[t]
    \centering
    \begin{minipage}{\linewidth}
        \subcaptionbox{Success and failure cases.\label{fig:demo_ours}}{
            \includegraphics[width=\linewidth]{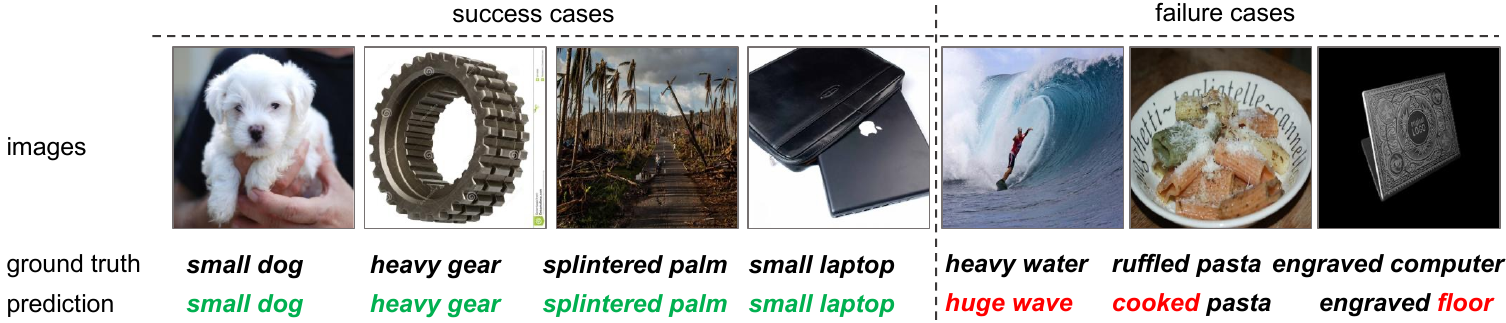}
        }
    \end{minipage}
    \begin{minipage}{\linewidth}
        \subcaptionbox{Comparison with model without LID.\label{fig:demo_compare}}{
            \includegraphics[width=\linewidth]{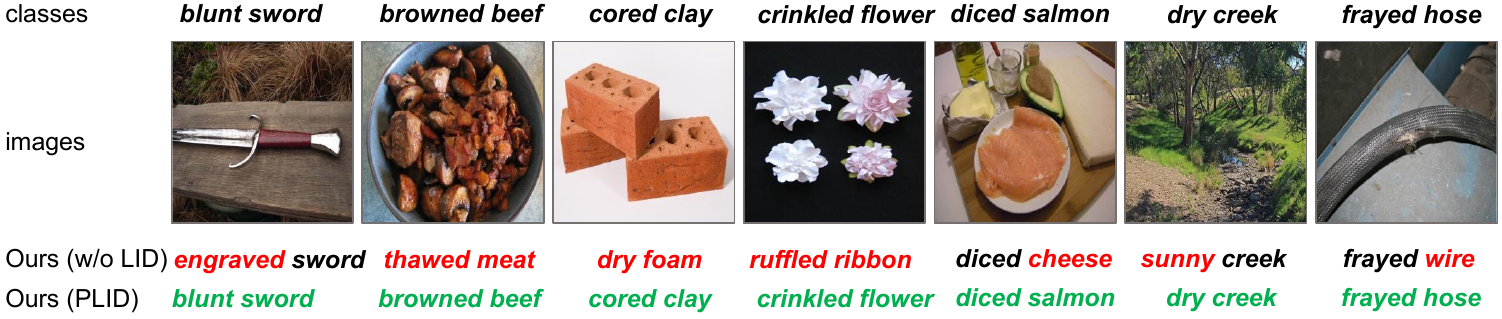}
        }
    \end{minipage}
    \caption{\footnotesize Case studies. In~\cref{fig:demo_ours}, we show the cases from MiT-States dataset that our method succeeds or fails. In~\cref{fig:demo_compare}, we compare the proposed method with and without language-informed distribution (LID) modeling. Correct predictions are in \textbf{\textcolor{JungleGreen}{green}} color, while incorrect predictions on state or object part are marked in \textbf{\textcolor{red}{red}}. }
    \label{fig:demo}
\end{figure}

\textbf{Class Distributions.} We use the tSNE to visualize the generated text embeddings $\mathbf{D}$ and the learned DSP from or $\mathbb{PLID}$ model in \cref{fig:tsne_all}, where the same set of 10 compositional (or state/object) classes are randomly selected from MIT-States dataset. It shows that by learning the distribution of each composition, state, and object from LLM-generated texts using Eq.~(\ref{eq:loss}) and~(\ref{eq:loss_so}) and TFE module, class embeddings can be distributed more compactly in each class (small intra-class variance), and better separated among multiple classes (large inter-class distance). This clearly shows why our proposed language-informed distribution modeling works in the CZSL task. 

\textbf{Case Study.} In \cref{fig:demo_ours}, we show some success and failure cases of our $\mathbb{PLID}$ model. For example, the \texttt{heavy water} case indicates an incorrect label while $\mathbb{PLID}$ could correctly predict it as \texttt{huge wave}. This shows the robustness of $\mathbb{PLID}$ against noisy labels. The last two failure cases reveal $\mathbb{PLID}$ still could make mistakes on the state prediction (\texttt{\underline{cooked} pasta}) and object prediction (\texttt{engraved \underline{floor}}), which indicates there are still rooms for improvement. In \cref{fig:demo_compare}, we show that $\mathbb{PLID}$ could work much better than the model without LID. For example, the \texttt{\underline{sunny} creek} and \texttt{frayed \underline{wire}} are incorrect potentially due to the lack of handling (\textit{i}) intra-class variety, as the \texttt{dry creek} images can be \texttt{sunny} and the \texttt{frayed hose} class could contain \texttt{wire} images, and (\textit{ii}) inter-class correlation, as the \texttt{sunny} (or \texttt{wire}) is correlated to both the \texttt{dry creek} images (or \texttt{frayed hose} images) and other \texttt{sunny} images (or \texttt{wire} images). 

\section{Conclusion}

In this work, we propose a novel CLIP-based compositional zero-shot learning (CZSL) method named $\mathbb{PLID}$. It leverages the generated text description of each class from large language models to formulate the class-specific Gaussian distributions. By softly prompting these language-informed distributions, $\mathbb{PLID}$ could achieve diversified and informative class embeddings for fine-grained compositional classes. Besides, we decompose the visual embeddings of image data into simple primitives that contain the basic states and objects, from which the re-composed predictions are derived to calibrate the prediction by our proposed stochastic logit mixup strategy. Experimental results show the superiority of the $\mathbb{PLID}$ method to prior arts on all common CZSL datasets. 


\section*{Acknowledgements} WT Bao and Y Kong were partially supported by the Office of Naval Research (ONR) grant N00014-23-1-2046 and N00014-23-1-2417. LC Chen and H Huang were partially supported by NSF IIS 2347592, 2347604, 2348159, 2348169, DBI 2405416, CCF 2348306, CNS 2347617.
Any opinions, findings, conclusions, or recommendations expressed in this material are those of the authors and do not necessarily reflect the views of ONR or NSF.

%
%
\bibliographystyle{splncs04}
\bibliography{main}

\clearpage
\begin{appendix}

\section*{Appendix}

\section{Broader Impact and Limitations} 

\textbf{Broader Impact.} The method in this work can be broadly extended to more multi-modality applications, such as general zero-shot learning, cross-modality compositional retrieval and generation, etc. Besides, the central idea of LLM-based modality alignment is not limited to text and image, but any modality that could reveal the semantic categories in practice is promising to explore in the future. The potential negative societal impact is that, the developers should be cautious by carefully examining the societal biases indicated by the generated class descriptions, though the LLMs we used are publicly accessible.

\vspace{2mm}
\noindent\textbf{Limitations.} One limitation is that the primitive decomposition could be difficult to learn when the states are non-visual concepts like \texttt{smelly}, \texttt{hot}, etc., even by the pre-trained CLIP model. Another limitation is that the generated descriptions by LLMs are not grounded to the image such that some distraction from generated descriptions could be introduced.

\section{Generating Compositional Class Descriptions}
\label{apd:llm_gen}

In this work, we choose T5-base, OPT-1.3B, GPT-3.5, and Mistral-7B models as the LLMs for compositional class description generation. For the T5 model, we follow the same setting as \cite{he2022synthetic} that uses the T5-base model for word-to-sentence generation. The T5-base model was pre-trained on the Colossal Clean Crawled Corpus dataset~\cite{raffel2020exploring} and finetuned on the CommonGen dataset~\cite{commongen}. Take the \texttt{painted ceiling} as an example, the results from T5-base model are:

\begin{adjustwidth}{0.5cm}{0.5cm}
\begin{lstlisting}[breakatwhitespace=true]
- A very old but beautifully decorated ceiling.
- A remodeled interior with a painted ceiling.
- A painted ceiling at a restaurant.
- Stained glass windows and a carved pattern on the ceiling.
- Painted ceilings and a fireplace.
- This apartment has a painted ceiling.
- A painted ceiling was a huge hit.
- A chandelier is painted in the middle of the ceiling.
- A stained glass window in a bathroom with a white painted ceiling.
- The ceiling of a hotel is painted.
- ...
\end{lstlisting}
\end{adjustwidth}

For the OPT model, we adopt the pre-trained OPT-1.3B model which is small but still could generate sentences of sufficiently good quality. Considering that it is a decoder-only LLM that relies on a few examples as the prompt for word-to-sentence generation, we design the prompt template with 4 examples as shown below.
\begin{adjustwidth}{0.5cm}{0.5cm}
\begin{lstlisting}[breakatwhitespace=true]
Keywords: sliced, potato, picture
Output: The picture features a beautifully arranged plate of thinly sliced potatoes.
###
Keywords: red, apple, photo
Output: In the photo, a bright red apple is the central focus, captured in stunning detail.
###
Keywords: leather, shoes, image
Output: The image showcases a sleek pair of leather shoes, meticulously designed to impress.
###
Keywords: broken, car, photo
Output: The photo captures the aftermath of a car accident with a broken vehicle lying on the road. 
###
\end{lstlisting}
\end{adjustwidth}
where the \texttt{Keywords} is followed by the words of the state, object, and the word randomly picked from the set $\{\texttt{photo}, \texttt{image}, \texttt{picture}\}$. Using the same example \emph{painted ceiling} as T5-base model, the generated sentences are:

\begin{adjustwidth}{0.5cm}{0.5cm}
\begin{lstlisting}[breakatwhitespace=true]
- The painting of the ceiling features an intricate pattern of intricate gold-colored paint and is framed by a white background.
- The ceiling has been painted with the pictures of these three characters, all arranged together.
- In the picture, the ceiling is covered in bright, colorful paintwork that has been painted on by hands painted white. The colors have been selected carefully.
- In the picture, the ceiling features painted decoration. The decoration resembles the surface of the sea, and has been painted in shades of blue. 
- The photograph captures both the bright colors of the painting atop the ceiling and the subtle shades of light reflecting off of it. 
- The large picture shows a large pattern painted onto the ceiling. The blue line shows paint dripping down.
- The wall behind the picture shows three different painted ceilings, in bright contrasting colors. A vibrant sky and blue skies are depicted against the dark brick wall.
- The ceiling of the room depicted in the painting could very well be painted in a few hours. The details of each object are clearly defined in its placement and position.
- Another photo of the same scene, this time featuring a ceiling painted in a stunning, white color. 
- A painted ceiling is shown, painted according to a specific design. this is a typical design that can also include decorative or functional elements. 
- ...
\end{lstlisting}
\end{adjustwidth}

It is clear that the generated class descriptions are much more diverse and informative than those of the OPT model.

\section{Covariance Sharing}
\label{apd:cov_share}

For the CZSL task, the spatial complexity of computing the covariance matrix $\boldsymbol{\Sigma}_{1:C}$ is $O(\lvert C^{(s)}\rvert^2d)$ which could be too heavy to compute if the number of the compositions is too large. For example, the C-GQA dataset contains 278K seen compositions which result in around $6\times 10^{13}$ floating elements of $\boldsymbol{\Sigma}_{1:C}$ for 768-dim text features. To handle this issue, we instead implement the $\boldsymbol{\Sigma}_{1:C}$ by sharing the covariance across attributes given the same object. This implies that the model is encouraged to learn the object-level distributions. 

Specifically, similar to the VLPD module of the main paper, we compute the mean $\boldsymbol{\mu}_{1:\lvert\mathcal{O}\rvert}$ and covariance $\boldsymbol{\Sigma}_{1:\lvert\mathcal{O}\rvert}$ over the objects by grouping $\mathbf{t}_y$ and $\mathbf{D}^{(y)}$ with object labels:
\begin{equation}
     \mathbf{t}_o = \frac{1}{\lvert\mathcal{Y}_o\rvert} \sum_{y\in \mathcal{Y}_o} \mathbf{t}_y, \;\;\;\; \mathbf{D}^{(o)} = \frac{1}{\lvert\mathcal{Y}_o\rvert} \sum_{y\in \mathcal{Y}_o} \mathbf{D}^{(y)},
\label{eq:toDo}
\end{equation}
where $\mathcal{Y}_o$ is the subset of compositions in $\mathcal{Y}$ that contains the same object as $y$. Then, all the pairwise margins $\mathbf{H}_o^{(m)}\in\mathbb{R}^{\lvert\mathcal{O}\rvert\times \lvert\mathcal{O}\rvert}$ in object space can be mapped back to $\mathbf{H}^{(m)}\in\mathbb{R}^{C\times C}$ in a compositional space by sharing it with all compositions in $\mathcal{Y}_o$. This could significantly reduce the computation load of the covariance while compromising the accuracy of distribution modeling.

Since the distribution modeling for both our $\mathbb{PLID}$ and ProDA is not applicable to the C-GQA dataset, we use the MIT States dataset to show the negative impact of sharing the covariance (see \cref{tab:sharecov}). It shows that the covariance sharing can significantly save the GPU memory (17.6 vs 32.5 GB), while still performing much better than ProDA.

\begin{table}[t]
    \centering
    \scriptsize
    \setlength{\tabcolsep}{0.8mm}
    \begin{tabular}{lclcclcc}
      \toprule
      Variants & Mem.(GB) && $\text{H}_\text{cw}$ & $\text{AUC}_\text{cw}$  && $\text{H}_\text{ow}$ & $\text{AUC}_\text{ow}$ \\ 
      \midrule
      ProDA~\cite{lu2022prompt} & 32.5   && 32.71 & 16.11     && 17.30  & 5.11  \\
      PLID (w. ShareCov)  & \textbf{17.6} && 38.50 \textcolor{red}{(-0.47\%)}  & 21.69 \textcolor{red}{(-0.43\%)}  && 19.81 \textcolor{red}{(-0.60\%)} & 7.04 \textcolor{red}{(-0.30\%)} \\
      \cellcolor{Gray}\textbf{$\mathbb{PLID}$} (full) &\cellcolor{Gray}22.2 &\cellcolor{Gray}&\cellcolor{Gray}\textbf{38.97} &\cellcolor{Gray}\textbf{22.12} &\cellcolor{Gray}&\cellcolor{Gray}\textbf{20.41}&\cellcolor{Gray}\textbf{7.34} \\
      \bottomrule
    \end{tabular}
    \caption{\footnotesize Effect of covariance sharing on MIT-States dataset. All methods use the same batch size of 64 for a fair comparison of GPU memory.}
    \label{tab:sharecov}
\end{table}

\section{Primitive-level Gaussian Modeling}
\label{sec:gauss_primitive}

To formulate the Gaussian distributions over the state classes and the object classes, we group the text embeddings of composition descriptions $\mathbf{D}$ by Eq.~(\ref{eq:toDo}), resulting in the distribution support points (DSP) $\mathbf{t}_o+\mathbf{D}^{(o)}$ and $\mathbf{t}_s+\mathbf{D}^{(s)}$ for a given object class $o$ and state class $s$, respectively. The DSPs are assumed to follow the state distribution $\mathcal{N}(\mathbf{t}_s,\boldsymbol{\Sigma}_s)$ or the object distribution $\mathcal{N}(\mathbf{t}_o,\boldsymbol{\Sigma}_o)$, where the covariances $\boldsymbol{\Sigma}_s$ and $\boldsymbol{\Sigma}_o$ are determined by $\mathbf{D}^{(s)}$ and $\mathbf{D}^{(o)}$, respectively.

Eventually, given the decomposed state visual features $f_s(\mathbf{v})$ and object visual features $f_o(\mathbf{v})$, the logit margin terms are defined as
\begin{equation}
    h_{k,s}^{(m)} = f_s(\mathbf{v})^{\top} \mathbf{A}_{k,s} f_s(\mathbf{v}), \;\;\; \text{and}\;\;\; h_{k,o}^{(m)} = f_o(\mathbf{v})^{\top} \mathbf{A}_{k,o} f_o(\mathbf{v}),
\end{equation}
where the index $k$ ranges within $[1,\lvert \mathcal{S}\rvert]$ for computing the state classification loss $\mathcal{L}_s$, and ranges within $[1,\lvert \mathcal{O}\rvert]$ for computing the object classification loss $\mathcal{L}_o$, respectively.

\section{More Implementation Details and Results}
\label{apd:impl}

\textbf{Implementation.} 
The training hyperparameters of our final model on each dataset are listed in \cref{tab:params}.

\begin{table}[t]
    \centering
    \setlength{\extrarowheight}{1mm}
    \begin{tabular}{l|ccc}
        \toprule
         Hyperparameters & MiT-States & UT-Zappos & C-GQA \\
         \hline
         max epochs & 20  & 25 & 20 \\
         base learning rate & 0.00005  & 0.0001 & 0.00001 \\
         weight decay & 0.00002  & 0.00001 & 0.00001 \\
         number of text descriptions & 64 & 32 & 64 \\
         number of image views & 8  &  8  & 8 \\
         attention dropout  &  0.5  & 0.1  & 0.1  \\
         weights of primitive loss & 0.1 & 0.01 & 0.01 \\ 
         \bottomrule
    \end{tabular}
    \caption{Hyperparameters of model implementation.}
    \label{tab:params}
\end{table}

\textbf{More Ablation Analysis.} In Table~\ref{tab:addexp}, we show more ablation study results on the design choices of our model. The first is to answer: \emph{Should we learn both the compositional and primitive feature space?} This is interesting because if the primitive space can be learned by the proposed VLPD, intuitively the original compositional space is redundant. In the first line of Table~\ref{tab:addexp}, we show that if we remove the compositional space but only learn primitive space to recompose, the performance experiences a large drop in all metrics. This can be explained by the intuition that, without a direct compositional recognition, the merits of \emph{explicitly} learned separatability and \emph{implicitly} learned compositionality will be totally lost. These are the keys to the success of the pioneering CZSL method CSP~\cite{csp_iclr23}. 

Besides, in Table~\ref{tab:addexp} line 2, we investigate whether the soft prompt is still useful or not based on our model, though it has been validated in prior CZSL literature~\cite{dfsp_cvpr23}. It shows that without the soft prompt, the performance decreases but not too much. However, it is still necessary as it drives the LLM text distributions to align with visual features in training.

Lastly, in Table~\ref{tab:addexp} lines 3-5, we further analyze the impact of TFE and VFE modules if they are implemented with the three-layer cross-attention Transformers. The two modules still show contributions to the performance gain. Moreover, compared to the default one-layer setting, using more Transformer layers does not improve the performance, even performing worse. 

\begin{table}[t]
    \centering
    \setlength{\tabcolsep}{3.5mm}
    \setlength{\extrarowheight}{1mm}
    \captionsetup{aboveskip=-\normalbaselineskip}
    \caption{More ablation study results.}
    \begin{tabular}{l|l|cccc}
    \toprule
    \multicolumn{2}{l|}{model variants} & $\text{H}_\text{cw}$ & $\text{AUC}_\text{cw}$  & $\text{H}_\text{ow}$ & $\text{AUC}_\text{ow}$ \\ 
    \midrule
         \multicolumn{2}{l|}{recompose only} & 30.02   & 13.88   & 15.46   &  4.35  \\
         \multicolumn{2}{l|}{w/o soft prompt} & 38.57   & 21.67   & 20.00   & 7.17   \\
         \hline
         \multirow{3}{*}{3-layers FE} & 
         TFE only &  36.89   &  19.93  & 18.77   & 6.42   \\
         & VFE only &  36.55   & 19.80   & 19.06   &  6.51  \\
         & TFE+VFE & 37.46  & 20.65  & 19.15  & 6.70   \\
         \hline
         \multicolumn{2}{l|}{full model} & 38.97 & 22.12 & 20.41 & 7.34\\
    \bottomrule
    \end{tabular}
    \label{tab:addexp}
\end{table}

\end{appendix}

\end{document}